\documentclass[10pt,conference,a4paper,conference]{IEEEtran}

\usepackage{cite}
\usepackage{amsmath,amssymb,amsfonts}
\usepackage[colorlinks=true, pagebackref]{hyperref}  %pagebackref hidelinks hyperref still needs to be put at the end!
\usepackage[square,sort&compress,comma,numbers]{natbib}% space in references ch (sort&compress for [1-4])
\usepackage{graphicx}
\usepackage{textcomp}
\usepackage{xcolor}
\usepackage{color,soul}
\usepackage[normalem]{ulem}
\usepackage{authblk}
\usepackage{diagbox}
\usepackage{pdfcomment}
\usepackage{todonotes}
\usepackage{multicol}

\def\BibTeX{{\rm B\kern-.05em{\sc i\kern-.025em b}\kern-.08em
		T\kern-.1667em\lower.7ex\hbox{E}\kern-.125emX}}

\newcommand\redst{\bgroup\markoverwith{\textcolor{red}{\rule[0.5ex]{2pt}{1pt}}}\ULon}

\begin{document}

	\title{Multi-Model Learning for Real-Time Automotive Semantic Foggy Scene Understanding \\via Domain Adaptation}

\author[1, 2]{Naif Alshammari}
\author[1, 3]{Samet Ak\c{c}ay}
\author[1, 4]{Toby P. Breckon\vspace{-.15cm}}

\affil[1]{ 
	Department of Computer Science, Durham University, Durham, UK
}%\vspace{1.5ex}}

\affil[2]{
	Department of Natural and Applied Science, Majma\textquotesingle ah University, Majma\textquotesingle ah, KSA}
\affil[3]{COSMONiO, Durham, UK}
\affil[4]{
	Department of Engineering, Durham University, Durham, UK \authorcr 
	{\tt\small
		\{\href{mailto:naif.alshammari@durham.ac.uk}{naif.alshammari},
		\href{mailto:samet.akcay@durham.ac.uk}{samet.akcay},
		\href{mailto:toby.breckon@durham.ac.uk}{toby.breckon}\}@durham.ac.uk
	}	
}

\maketitle

\begin{abstract}
Robust semantic scene segmentation for automotive applications is a challenging problem in two key aspects: (1) labelling every individual scene pixel and (2) performing this task under unstable weather and illumination changes (e.g., foggy weather), which results in poor outdoor scene visibility. Such visibility limitations lead to non-optimal performance of generalised deep convolutional neural network-based semantic scene segmentation. In this paper, we propose an efficient end-to-end automotive semantic scene understanding approach that is robust to foggy weather conditions. As an end-to-end pipeline, our proposed approach provides: (1) the transformation of imagery from foggy to clear weather conditions using a domain transfer approach (correcting for poor visibility) and (2) semantically segmenting the scene using a competitive encoder-decoder architecture with low computational complexity (enabling real-time performance). Our approach incorporates RGB colour, depth and luminance images via distinct encoders with dense connectivity and features fusion to effectively exploit information from different inputs, which contributes to an optimal feature representation within the overall model. Using this architectural formulation with dense skip connections, our model achieves comparable performance to contemporary approaches at a fraction of the overall model complexity.  
\end{abstract}

%with dense connectivity, skip connections and features fusion techniques
%We evaluate our approach on challenging foggy datasets, including synthetic (\textit{Foggy Cityscapes}) and real-world (\textit{Foggy Zurich} and \textit{Foggy Driving}) examples

	\section{Introduction}
Semantic scene segmentation is an active research topic that targets robust pixel-level image classification. However, as the reported performance of many state-of-the-art scene understanding algorithms is limited to ideal weather conditions, extreme weather and illumination variation could lead to unexpectedly inaccurate scene classification and segmentation \cite{finlayson2009entropy, maddern2014illumination, alvarez2011road, upcroft2014lighting, krajnik2015visual, corke2013dealing}. To date, too little attention has been paid to address the issue of automotive scene understanding under extreme weather conditions (e.g., Foggy weather) \cite{foggy18, foggy19}, as the multitude of proposed deep learning approaches are generally only evaluated on ideal weather conditions. To overcome this shortcoming, the present paper introduces an efficient algorithm that tackles the challenge of automotive scene understanding in extreme weather conditions using a novel multi-modal learning approach that translates foggy scene images to clear scenes and utilizes both depth and luminance information to achieve superior semantic segmentation performance.  

\begin{figure}[t!]
	
	\begin{minipage}[b]{1.0\linewidth}
		\centering
		\centerline{\includegraphics{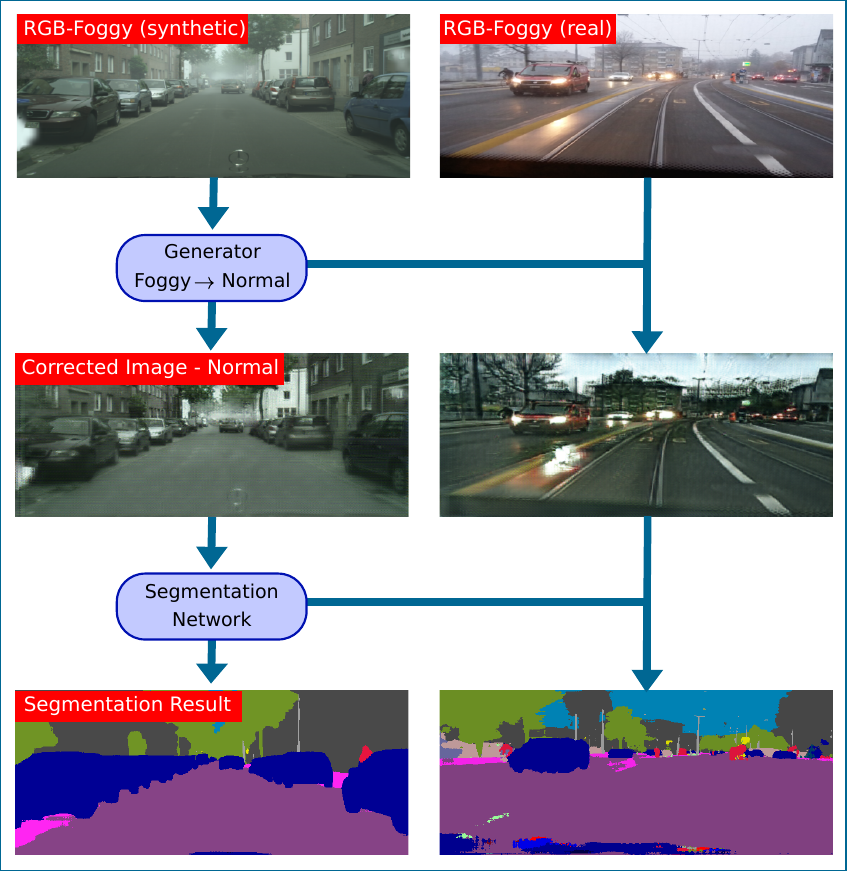}}
		
	\end{minipage}
	\vspace{-.55cm}
	\caption{An illustration of our semantic segmentation approach under foggy weather conditions, trained on \textit{Foggy Cityscapes} \cite{foggy18} (partially synthetic data) and evaluated on \textit{Foggy Driving} \cite{foggy18} (real foggy scenes). Degraded scenes visibility present under foggy weather conditions are corrected using domain adaptation \cite{CycleGAN2017} to serve a better semantic segmentation performance.} %, such technique will produce optimal inputs to semantic segmentation.}   
	\label{fig:intro}
	\vspace{-15px}
\end{figure}

\begin{figure*}[t!]
	\begin{minipage}[b]{1.0\linewidth}
		\centering
		\centerline{\includegraphics[width=\columnwidth]{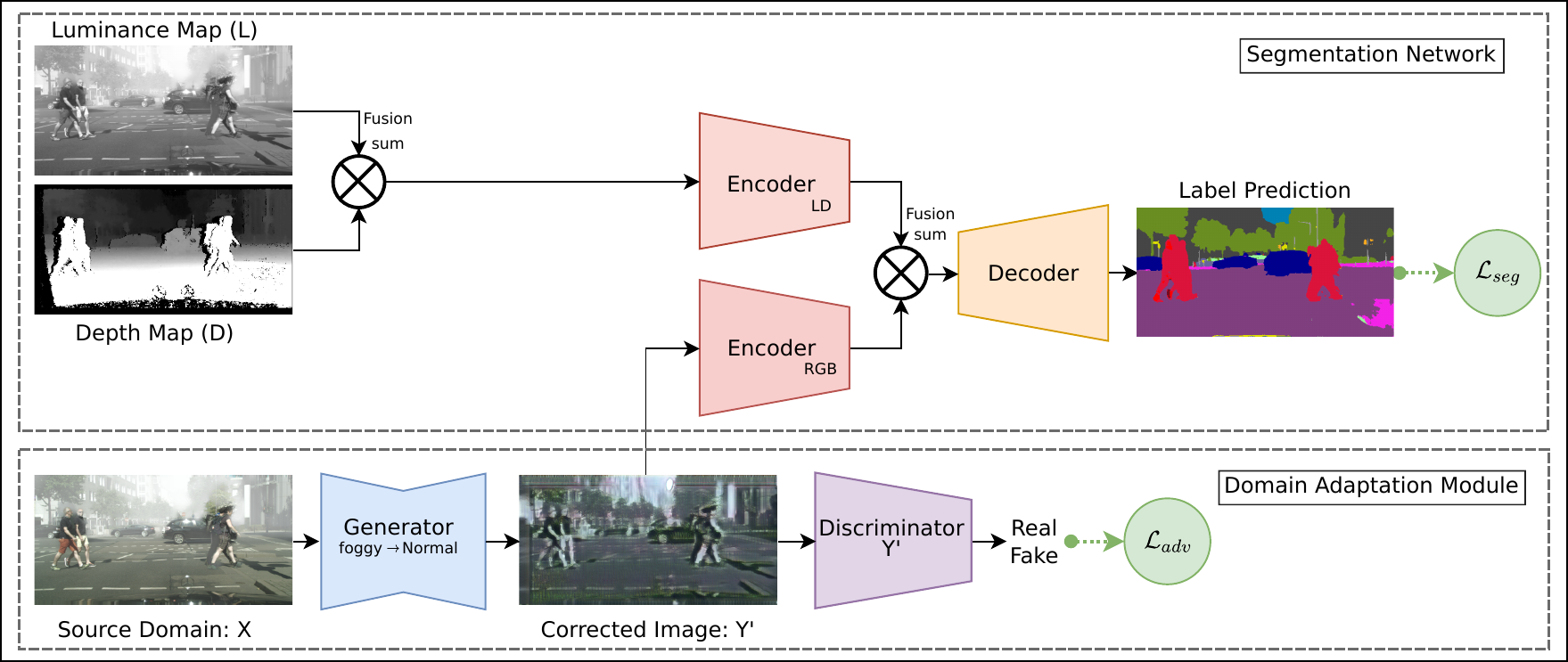}} %myModel3
	\end{minipage}
	%\vspace{-.55cm}
	\caption{Overview of our approach using \cite{CycleGAN2017, ldfnet}. The source domain $X$ (foggy scene) mapped to the target domain $Y'$ (corrected image). Subsequently, the corrected image $Y'$ is fed to the RGB encoder in the semantic segmentation network. The Depth (D) and luminance (L) images are incorporating RGB colour via the LD encoder. Finally, the output from the two encoders is passed to the semantic segmentation decoder.}
	%\todo[inline]{Target Domain picture doesn't look realistic. Can we find better one?}
	\label{fig:arch}
	\vspace{-15px}
\end{figure*}

Previous attempts to tackle the issue of scene understanding under non-ideal weather conditions for shadow removal and illumination reduction \cite{maddern2014illumination, alvarez2011road, upcroft2014lighting, krajnik2015visual}, haze removal and scene defogging \cite{defog1, defog2, defog3}, and foggy scene understanding \cite{foggy18, foggy19} are mostly based on conventional image enhancement and dehazing methods. Despite the general trend of performance improvement within automotive scene understanding \cite{pyramid, deeplab, resnet, refineNet17}, there is still significant room for improvement across the spectrum of non-ideal operating conditions. In parallel with using recent image segmentation techniques \cite{densenet, fusenet, skip3, ldfnet}, employing the concept of image-to-image translation to map one domain onto another \cite{CycleGAN2017, pix2pix2016} is a useful step that enables accurate semantic segmentation performance under extreme weather conditions.

In this work, we propose an efficient end-to-end automotive semantic scene understanding capable of performing under foggy weather conditions. We employ domain adaptation within scene understanding as a method to correct for the degraded visibility present under foggy weather conditions. In addition, we propose a lightweight semantic segmentation architecture that incorporates RBG colour, luminance and depth images via distinctive encoders contributing to a deeper extraction for the representation of different features, which leads to superior segmentation performance. As an integration methodology within encoders, we use a fusion-based connection. To avoid information loss and share high-resolution features in the latter reconstruction stages of CNN upsampling, we leverage skip-connections \cite{unet, skip1, skip2, skip3}. 
%\vspace{-7px}

	\section{Related Work}
\label{sec:relatedwork}
The related work is organized into two main categories: (i) semantic segmentation (Section \ref{sec:seg}) and (ii) domain transfer (Section \ref{sec:style}). %, and (3) illumination-invariant and perceptual colour space computation (Section \ref{sec:illum}).

\subsection{Semantic Segmentation}
\label{sec:seg}
Modern segmentation techniques utilize deep convolutional neural networks and outperform the traditional approaches by a large margin \cite{segnet17, deeplab, resnet, pyramid, refineNet17}. These contributions use a large dataset such as ImageNet \cite{imagenet} for pre-trained models. Recent segmentation techniques have distinct characteristics denoted by their design such as: (1) network topology: pooling indices \cite{segnet17}, skip connection \cite{unet}, multi-path refinement \cite{refineNet17}, pyramid pooling \cite{pyramid}, fusion-based architecture \cite{fusenet} and dense connectivity \cite{densenet}, (2) varying input: colour RGB or RGB-D with depth \cite{fusenet, holder_depth}, depth and luminance \cite{ldfnet}, and illumination invariance \cite{naif}, and (3) consideration of adverse-weather conditions \cite{foggy18, foggy19, foggy_pure}. As the main objective of this work is semantic scene segmentation under foggy weather conditions, recent studies in this specific domain are specifically presented in this section. %\vspace{-.1cm}

Different approaches have been proposed for tackling the issue of scene understanding under adverse weather conditions. To address illumination changes, an illumination-invariant colour space approach was proposed in \cite{krajnik2015visual, kim2017pca, naif} to minimize scene colour variations due to varying scene lighting conditions. Other approaches \cite{foggy_old, foggy18, foggy19} addressed scene segmentation under foggy weather conditions using a semi-supervised approaches through generating synthetic foggy images from real-world data, and augmenting clear images with their synthetic fog images. By adapting segmentation models from day to night scenes, \cite{sakaridis2019guided} addressed the issue of vision under nocturnal conditions. 

Similarly, our model is trained on \textit{foggy} scenes (synthetic images) adapted to \textit{normal} using domain adaptation via style transfer (Section \ref{sec:style}) . Inspired by LDFNet \cite{ldfnet}, we employ the idea of incorporating luminance and depth alongside RGB, utilizing skip connections as well as fused features to perform semantic segmentation under foggy weather conditions. 

\subsection{Domain Transfer}
\label{sec:style}
Transferring an image from its real domain to another differing domain allows multiple uses of such images taken in complex environments or generated in different forms. Using recent advances in the field of image style transfer, \cite{style_first}, where target images are generated by capturing the style texture information of the input image by utilizing the Gram matrix, work by \cite{amir_refer} shows that image style transfer (from the source domain to the target domain) is the fundamental process by which the differences between source and target distribution are minimized.  

Recent methods \cite{pix2pix2016, CycleGAN2017, sketch} used Generative Adversarial Networks (GAN) \cite{gan} to learn mapping from source to target images. Based on training over a large dataset for specific image style,  CycleGAN \cite{CycleGAN2017} shows an efficient approach to transfer image style from one image domain into another.  

Within the context of semantic segmentation, we take advantage of CycleGAN \cite{CycleGAN2017} to improve semantic segmentation by generating target scenes (clear-weather scenes) from the source domain (foggy scenes images) as source images $I_{x}$ mapped into a target domain $I_{y}$ --- hence significantly increasing our available image data training resources.
	\section{Proposed Approach}
\label{sec:method}
Our main objective is to train an end-to-end network that semantically labels every pixel in a scene that is invariant to both weather and illumination variations. We make use of \textit{Foggy Cityscapes} dataset \cite{foggy18} for training. However, as the visibility is degraded due to fog, we attempt to reduce this sensing challenge using a model trained to transfer the style of \textit{foggy} scenes to \textit{normal}. \textit{Foggy Driving} \cite{foggy18} and \textit{Foggy Zurich} \cite{foggy19} are used as independent test sets comprising real world evaluation examples. 

In general, our approach consists of two sub-components, namely domain transfer and semantic segmentation (each functioning as an integrated unit). These sub-components produce two separate outputs: a transformed clear-scene image (generated from a foggy domain) and semantic pixel labels. The pipeline of our approach is shown in Figure \ref{fig:arch}. In this section, we provide a detailed overview of these two sub-components: Domain Transfer (Section \ref{sec:transfer}) and Semantic Segmentation (Section \ref{sec:semeseg}).\vspace{-3px} %(1) the network architecture we developed, 

\subsection{Domain Transfer}
\label{sec:transfer}
Our goal is to learn mapping $\mathcal{D:} X \rightarrow Y$ from the source domain $X$ (foggy scenes) to the target domain $Y$ (clear-weather) for which we assume that the scene visibility level in the constructed image is the optimal input to the subsequent Semantic Segmentation (Section \ref{sec:semeseg}). We use CycleGAN \cite{CycleGAN2017} to learn this domain transfer mapping function (shown in Figure \ref{fig:arch}, lower). A generator $G_{X \rightarrow Y}$ (generating clear scenes samples $Y'$) and a discriminator $D_{Y}$ (discriminating between $Y$ and $Y'$) are used to perform the mapping function from the source and target domains. The loss for each generator $G$ coupled with a discriminator $D$ is calculated as follows:  

% GAN loss
\begin{equation}\label{eq:adv_xy}
	\begin{aligned}
	\mathcal{L}_{adv}(X \rightarrow Y) = \min_{G_{Y \rightarrow X}} \max_{D_{Y}} \mathbb{E}_{y \sim \mathbb{P}_{d}(y)} [\log(D)_{(y)}]+ \\ \mathbb{E}_{x \sim \mathbb{P}_{d}(x)} [\log(1-D_{Y}(G_{X \rightarrow Y}(x)))],
	\end{aligned}
\end{equation}where $\mathbb{P}_{d}$ is the data distribution, $X$ the source domain with samples $x$ and $Y$ the target domain with the samples $y$. 

\subsection{Semantic Segmentation}
\label{sec:semeseg} % adopted \cite{erfnet}, used fusion technique \cite{fusenet} and layers dense connections \cite{densenet}.
\begin{figure}
	\centering
	\includegraphics[width=\linewidth]{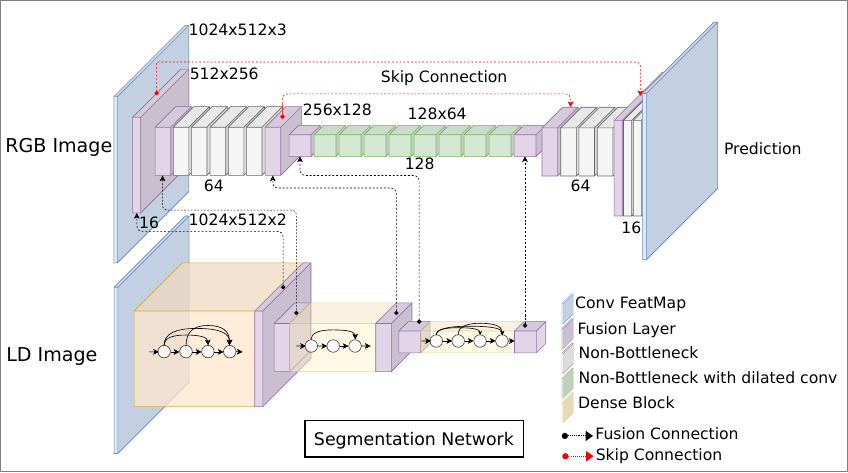}
	\caption{Details of the segmentation network which consists of two encoders taking two types of inputs: \textbf{RGB Image} and \textbf{LD Image} (Luminance and Depth channels).}
	\vspace{-20px}
	\label{fig:seg-net}
\end{figure}

As a subsequent component of the overall model, our pipeline performs the task of semantic segmentation on the corrected images $Y'$ (mapped from \textit{foggy} $X$ to \textit{normal} $Y$ weather conditions via domain adaptation \cite{CycleGAN2017} as $G_{X \rightarrow Y}(X)=Y'$) incorporated with luminance $L$ and/or depth $D$) (shown in Figure \ref{fig:arch}, upper). Motivated by \cite{ldfnet}, we use an auto-encoder architecture for semantic segmentation, consisting of two distinctive encoders for image downsampling and features extraction: RGB encoder ($E_{RGB}$) and luminance with depth encoder ($E_{LD}$) or only luminance ($E_{L}$)  if depth is not available (Figure \ref{fig:arch}, upper). Using explicit encoders for RGB colour, luminance and depth are considered to efficiently exploit information
from different inputs such as luminance and depth \cite{fusenet, ldfnet}. As seen in Figure \ref{fig:seg-net}, we utilize dense-connections and features fusion to gain an optimal feature representation. To subsequently upsample the feature maps representation to the original input dimension, corresponding decoder is used (Figure \ref{fig:arch}, upper). Our encoder-decoder architecture is leveraging skip-connections to keep sharing high-level features which leads to a superior semantic segmentation
performance (Figure \ref{fig:seg-net}). %In addition, dense-connections, features fusing and extracted fusion maps are implemented in the baseline architecture.

\vspace{.1cm}
\noindent\textbf{{RGB encoder:}}
\label{sec:rgb}
Designed to deal with a three-channel RGB input, the RGB encoder ($E_{RGB}$) comprises of three downsampler blocks with convolutional and max pooling layers followed by batch normalization and \textit{ReLu()} activation function (\{\textit{16, 64, 128}\}, respectively). Subsequently, five non-bottleneck modules are implemented including the factorized convolutions (convolution kernel $n$$\times$$n$ factorized into $n$$\times$$1$ and $1$$\times$$n$ kernels), each followed by batch normalization and \textit{ReLu()} with residual connections. With dilated and factorized convolutions, eight non-bottleneck modules were implemented as the last component of $E_{RGB}$ (Figure \ref{fig:seg-net}). 

\vspace{.1cm}
\noindent\textbf{{LD encoder:}}
\label{sec:d&l}  
Unlike the RGB encoder, the luminance and depth encoder ($E_{LD}$) deals with luminance and depth images (concatenated as two-channel input). As a parallel functioning to $E_{RGB}$, $E_{LD}$ is designed with a dense connectivity \cite{densenet} technique to enhance the information flow from the earlier to the last layers. Specifically, $E_{LD}$ consists of a downsampler (the same as in $E_{RGB}$) followed by three dense blocks with \{\textit{16, 64, 128}\} channels, respectively. Each dense block is followed by a transition layer designed with $1$$\times$$1$ and followed with $2$$\times$$2$ average pool layer. As some datasets do not contain depth maps, we used a luminance only encoder ($E_{L}$) that is identical to ($E_{LD}$) except that it only take luminance channel. We make use of a distinct encoder for luminance and depth to exploit deeper and better representation from the depth and luminance maps \cite{fusenet, ldfnet} (Figure \ref{fig:seg-net}).

$E_{RGB}$ and $E_{LD}$ are linked by fusing output layers from blocks sharing the same number of channels among $E_{RGB}$ and $E_{DL}$. Since it requires less computational cost, the fusion connectivity is implemented by summing the two layers such that for inputs $x$ and $y$, the fused feature map is $E_{RGB}(x) + E_{LD}(y)$ or $E_{L}(y)$.
\begin{figure}[t!]
	\begin{minipage}[b]{1.0\linewidth}
		\centering
		\centerline{\includegraphics{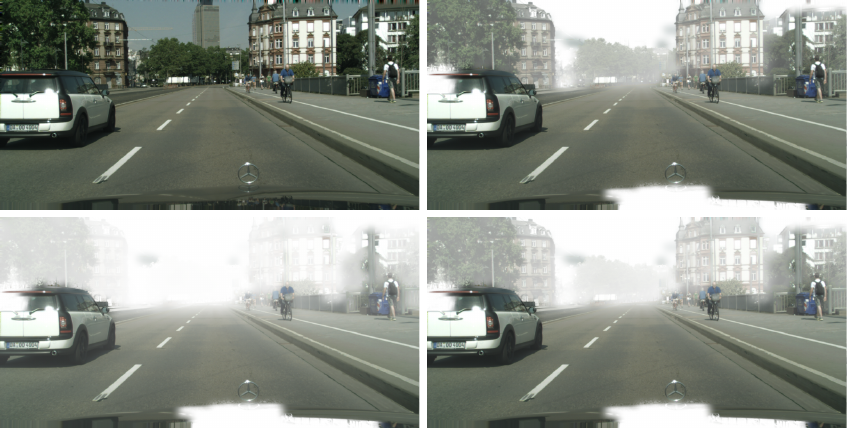}} %myModel3
	\end{minipage}
	\vspace{-.55cm}
	\caption{Sample image from \textit{Cityscapes} \cite{cityscapes} (top left) followed by (clockwise) foggy images (partially synthetic) with varying visibility (light to dense) from \textit{Foggy Cityscapes} \cite{foggy19}.}
	\label{fig:city_imgs}
	%\vspace{-14px}
\end{figure}

\vspace{.1cm}
\noindent\textbf{{Decoder:}}
\label{sec:dec}
After fusing the feature maps extracted from the last layer of $E_{RGB}$ and either $E_{DL}$ or $E_{L}$, a decoder upsamples the feature maps to the original resolution. The upsampling is implemented in three stages \{\textit{64, 16, 19}\}. In the first two stages, transposed convolution, batch normalization, and \textit{ReLu()} activation function, as well as two non-bottleneck modules, are employed. As the last component in the encoder, the transposed convolution layer maps the output to 19 class labels which we aim to predict (Figure \ref{fig:seg-net}).

\vspace{.1cm}
Unlike LDFNet \cite{ldfnet}, we utilize skip connections for the fused features from the encoders into the decoder to avoid the loss of the high-level spatial features after being downsampled (Figure \ref{fig:seg-net}). The fused feature maps \{64, 16\} passed from the encoders are concatenated with the corresponding upsampled feature maps in the decoder. As a semantic segmentation loss function, cross-entropy with pixel-wise \textit{softmax()} is used summing over all pixels within a patch as follows:

\begin{equation}\label{eq:seg1}
	\begin{aligned}
	{P}_{k}(x) =   \frac{e^{ak(x)}}{\sum_{k^\prime =1}^{K}e^{a_{k^\prime}(x)}},
	\end{aligned}
\end{equation}

\begin{equation}\label{eq:seg2}
	\begin{aligned}
	\mathcal{L}_{seg} = -\log(P_{l}(S(x))),
	\end{aligned}
\end{equation}where $S(x)$ denotes the output of the segmentation network, $K$ is the number of classes, $P_{k}(x)$ is the approximated maximum function, and $l$ is the ground truth label, $a_{k}(x)$ the feature activation for the channel $k$. As an overall loss, a joint overall loss function for our model is calculated as follows:

\begin{figure}[t!]
	\begin{minipage}[b]{1.0\linewidth}
		\centering
		\centerline{\includegraphics{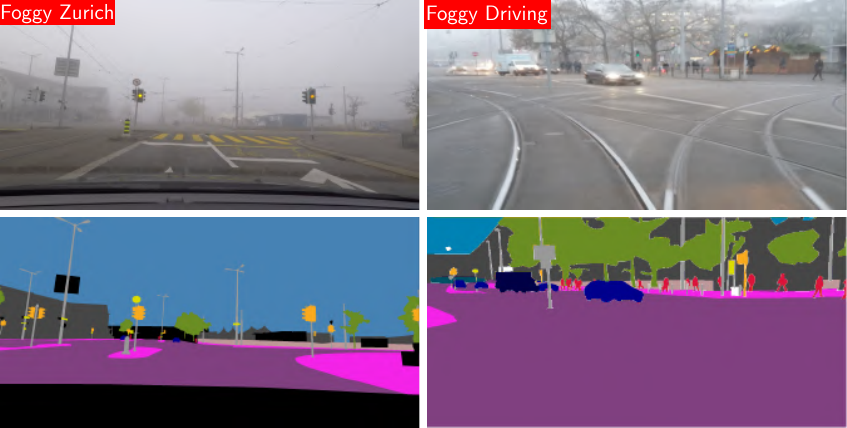}} %myModel3
	\end{minipage}
	\vspace{-.55cm}
	\caption{Sample images from \textbf{Foggy Zurich} \cite{foggy19} and \textbf{Foggy Driving} \cite{foggy18} (real-world foggy datasets) along with their annotations.}
	\label{fig:zurich_imgs}
	%\vspace{-14px}
\end{figure}

\begin{equation}\label{eq:joint_loss}
	\mathcal{L} = \lambda_{adv} \mathcal{L}_{adv}(I_{y}) +
	\lambda_{seg}\mathcal{L}_{seg}(I_{s}),
\end{equation}where $\lambda_{adv}$ and $\lambda_{seg}$ are weighting coefficients, and dynamically updated using the homoscedastic uncertainty technique to weight and balance the two losses \cite{cipolla2018multi}.
	\section{Dataset}
\label{sec:data}
The availability of numerous well-annotated datasets \cite {cityscapes, kitti, imagenet, pascal} has led to a proliferation of semantic segmentation studies. In this section, we will present the following datasets used in this paper: \textit{Cityscapes dataset} \cite{cityscapes} as the base dataset representing clear scenes, and \textit{Foggy Cityscapes dataset} \cite{foggy18} as a partially synthetic data where fog is added into \cite{cityscapes} (fine weather). As real-world datasets for \textit{foggy} weather conditions, \textit{Foggy Driving} \cite{foggy18} and \textit{Foggy Zurich} \cite{foggy19} are used.

\vspace{.1cm}
\noindent\textbf{{Cityscapes Dataset:}}
We evaluate our approach on the \textit{Cityscapes} \cite{cityscapes} (Figure \ref{fig:city_imgs}), a large dataset collected for urban-scene semantic segmentation. The dataset comprises of $2,975$ training and $500$ testing image examples (resolution: $1024\times2048$) with 19 pixel classes: \textit{\{road, sidewalk, building, wall, fence, pole, traffic light, traffic sign, vegetation, terrain, sky, person, rider, car, truck, bus, train, motorcycle and bicycle\}}. In addition to semantic labels, \textit{Cityscapes} provides disparity dataset labelled using Semi-Global Matching \cite{hirschmuller2007stereo}, used as a complementary information for semantic segmentation.

\vspace{.1cm}
\noindent\textbf{{Foggy Cityscapes Dataset:}}
\label{sec:foggy_city}
\textit{Foggy Cityscapes} \cite{foggy18} is a partially synthetic data generated from Cityscapes \cite{cityscapes} by adding synthetic fog to the real images using fog simulation \cite{foggy18}. Three different versions of this dataset (shown in Figure \ref{fig:city_imgs}) exist with varying fog density levels (controlled using attenuation coefficient $\beta \in \{0.005, 0.01, 0.02\}$ \textemdash from light to dense fog) and were used in the present study. This dataset inherits the annotations from \textit{Cityscapes} \cite{cityscapes} as labels for the synthetic foggy datasets as well as disparities. \textit{Foggy Cityscapes} dataset consists of $8925$ training and $1500$ testing image examples (resolution: $1024\times2048$).

\begin{table*}[t]
	\centering
	\resizebox{\linewidth}{!}{
		
		\begin{tabular}{c|c|c|c|c|c|c|c}
			%\hline
			\multicolumn{4}{c|}{Methods} & \multicolumn{1}{c|}{Foggy Zurich} & \multicolumn{1}{c|}{Foggy Driving} & \multicolumn{2}{c}{Complexity of the Network} \\ \hline \hline 
			{Models} & {Network Architecture} & Training & Fine-Tuning & {Mean IoU}  & {Mean IoU} & {Number of Parameters} & {FPS}\\ \hline \hline 
			CMDAda\cite{foggy19} & AdSegNet \cite{similar_naif} - DeepLab-v2 \cite{deeplab}      & C & \textemdash    & 25.0  & 29.7 & 44.0M  &20 \\ \hline
			SFSU \cite{foggy18} & Dilated Conv. Net. (DCN) \cite{dcn}  & C & FC (498)       & 35.7 & 46.3 & 134M  &- \\ \hline 
			CMAda2+ \cite{foggy_old}& RefineNet-ResNet-101 \cite{refineNet17}  & C & FC (498)      & 43.4  & 49.9 & 118M & 22\\ \hline 
			CMAda3+ \cite{foggy19}  & RefineNet-ResNet-101 \cite{refineNet17}  & C & FC (498)       & 46.8 &  49.8 & 118M & 22 \\ \hline
			
			Hanner \textit{et al.} \cite{foggy_pure} & RefineNet-ResNet-101 \cite{refineNet17}  & C & FS (24,500)         & 40.3 & 48.4 & 118M  & 22\\ \hline
			Hanner \textit{et al.} \cite{foggy_pure} & RefineNet-ResNet-101 \cite{refineNet17}  & C & FS (498)         & 42.7 &  48.6 & 118M  & 22\\ \hline
			Hanner \textit{et al.} \cite{foggy_pure} & RefineNet-ResNet-101 \cite{refineNet17}  & C & FC+FS (498)         & 41.4  & 50.7 & 118M  & 22\\ \hline
			
			Hanner \textit{et al.} \cite{foggy_pure} & BiSeNet \cite{bisenet}  & C & \textemdash     & 16.1 &  27.2 & 50.8M &-\\ \hline
			Hanner \textit{et al.} \cite{foggy_pure} & BiSeNet \cite{bisenet}  & C & FC (498)       & 25.0 &  30.3 & 50.8M &-\\ \hline
			Hanner \textit{et al.} \cite{foggy_pure} & BiSeNet \cite{bisenet}  & C & FS (24,500)    & 27.8 &  30.9 & 50.8M  &-\\ \hline
			Hanner \textit{et al.} \cite{foggy_pure} & BiSeNet \cite{bisenet}  & C & FS (498)       & 27.6 &  31.8 & 50.8M  &-\\ \hline
			Hanner \textit{et al.} \cite{foggy_pure} & BiSeNet \cite{refineNet17}  & C & FC+FS (498)       & 35.2 &  30.9 & 118M  & 22\\ \hline
			
			Ours w/o domain adaptation     & \textemdash & C & FC (498) &  8.7  & 17.6 & \textbf{2.4M} & \textbf{42} \\ \hline
			
			Ours w/ domain adaptation     & \textemdash & C & FC (498) &  21.4  & 29.4 & \textbf{13.8M} & 20 \\ \hline
			
	\end{tabular}}
	\vspace{3px}
	\caption{Quantitative comparison of semantic segmentation on Foggy Zurich \cite{foggy19} and Foggy Driving \cite{foggy18} datasets of our approach against state-of-the-art approaches. \textbf{C}: Cityscapes \cite{cityscapes}; \textbf{FC} Foggy Cityscapes \cite{foggy18}; \textbf{FS}: Foggy Synscapes \cite{foggy_pure}. The speed comparison (frames per second (fps) is based on the Cityscapes \cite{cityscapes} test dataset.} 

\label{tab:main_res}
%\vspace{-14px}
\end{table*}

\vspace{.1cm}
\noindent\textbf{{Foggy Driving Dataset:}}
The \textit{Foggy Driving} dataset \cite{foggy18} (Fig. \ref{fig:zurich_imgs}) is a real-world dataset collected in foggy-weather conditions, consisting of $101$ images (resolution: $960\times1280$) with annotations for semantic segmentation and object detection tasks. Following \textit{Cityscapes} \cite{cityscapes} dataset, the \textit{Foggy Driving} dataset is labelled with 19 classes (33 images with fine annotations and 68 images coarsely annotated).  

\vspace{.1cm}
\noindent\textbf{{Foggy Zurich Dataset:}}
The \textit{Foggy Zurich} \cite{foggy19}  (Fig. \ref{fig:zurich_imgs}) is a real-world foggy-scenes dataset consisting of $3808$ images (resolution: $1920\times1080$) collected in Zurich. Following the approach of \textit{Cityscapes} \cite{cityscapes}, \textit{Foggy Zurich} provides pixel-level annotations for 40 scenes (finely annotated), including dense fog.

	\section{Implementation Details}
\label{sec:implement}
We implement our approach in PyTorch \cite{pytorch}. For optimization, we employ ADAM \cite{adam} with an initial learning rate of $5\times10^{-3}$ and momentum of $\beta_1 = 0.5, \beta_2 = 0,999$. By following \cite{enet} and \cite{ldfnet}, we weight the classes of the dataset duo to imbalance number of pixels of each class in the dataset as follows:

\begin{equation}\label{eq:cweight}
	\begin{aligned} 
		\omega_{class}=\frac{1}{\ln(c+p_{class})},
	\end{aligned}
	\vspace{0.2cm} 
\end{equation} where c is an additional parameter empirically set to 1.10 to restrict the class weight and $p_{class}$ is the probability of belonging to that class. We train the model for $100$ epoch by using NVIDIA Titan X and GTX 1080Ti GPUs. We apply data augmentation in training using random horizontal flip for high resolution images ($256\times512$). For semantic accuracy evaluation, we use the following evaluation measures: class average accuracy, the mean of the predictive accuracy over all classes, global accuracy, which measures overall scene pixel classification accuracy, and mean intersection over union (mIoU).

For the \textit{Foggy Cityscapes} \cite{foggy18}, \textit{Foggy Driving} \cite{foggy18}, and \textit{Foggy Zurich} \cite{foggy19} datasets we train using the available information. For all datasets, as a complementary information source, we make use of the luminance transformation, which is a translated grayscale image $L$ derived from ${I}_{RGB} \in \{I_R, I_G, I_B\}$ to both reduce the noise and improve feature extraction, defined as follows: 

\begin{equation}\label{eq:Limg}
	\begin{aligned} 
		L=0.299(I_R) + 0.587(I_G) + 0.144(I_B)
	\end{aligned} 
\end{equation}

	\vspace{.1cm}
\section{Evaluation}
We evaluate the performance of our proposed approach on the benchmark foggy weather conditions datasets: \textit{Foggy Cityscapes} \cite{foggy18}, \textit{Foggy Driving} \cite{foggy18}, and \textit{Foggy Zurich} \cite{foggy19}. The evaluation was performed as follows:

\vspace{.1cm}
\renewcommand{\labelenumii}{\Roman{enumii}}
\begin{enumerate}
	\item \label{point1} We train the domain adaptation component (Section \ref{sec:transfer}) (employed later as a sub-component (Fig. \ref{fig:arch}) trained in step \ref{point3}) on the \textit{Cityscapes} dataset (\textit{normal} weather) \cite{cityscapes} and Foggy Cityscapes (\textit{adverse} weather) \cite{foggy18} to map from \textit{foggy} scenes to \textit{normal}. 
	
	\item \label{point2} In the same manner, we train the semantic segmentation component (Section \ref{sec:semeseg})  on the \textit{Cityscapes} dataset \cite{cityscapes} (\textit{normal} weather).
	
	\item \label{point3} Models obtained from steps (\ref{point1}, \ref{point2}) are fine-tuned within a unified architecture using \textit{refined Foggy Cityscapes} dataset \cite{foggy18} (a subset including 498 training and 52 testing better quality images).
	
	\item \label{point4} The fine-tuned architecture in step \ref{point3} is evaluated on \textit{Foggy Driving} \cite{foggy18} and \textit{Foggy Zurich} \cite{foggy19}. 
\end{enumerate}

\begin{figure*}[t!]
	\begin{minipage}[b]{1.0\linewidth}
		\centering
		\centerline{\includegraphics[width=\columnwidth]{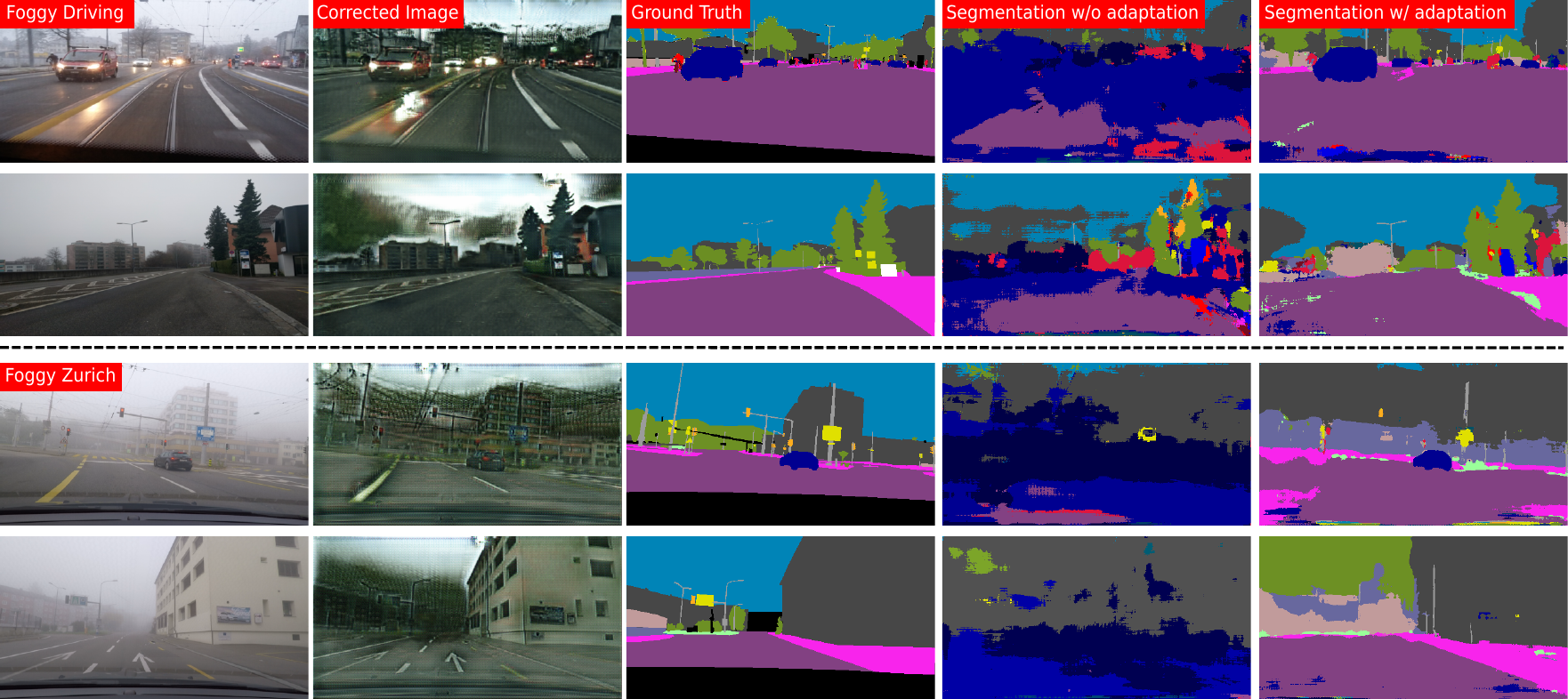}} %myModel3 16.0 w 13.3 h
	\end{minipage}
	\vspace{-.55cm}
	\caption{Semantic segmentation predictions on \textbf{Foggy Driving} \cite{foggy18} and \textbf{Foggy Zurich} \cite{foggy19} for the proposed approach. The left column shows two scenarios of each dataset followed by \textbf{Corrected Image:} corrected image using \textit{domain adaptation}; \textbf{Ground Truth:} ground truth segmentation; \textbf{Segmentation w/o adaptation:} segmentation results without using \textit{domain adaptation}; \textbf{Segmentation w/ adaptation:} segmentation results with \textit{domain adaptation}.}
	\label{fig:main_res}
	%\vspace{-14px}
\end{figure*}

\begin{table}[t!]
	\centering
	\begin{tabular}{c|c|c|c}
		\diagbox{Methods}{Results} & {Global avg.} & {Class avg.} & {Mean IoU} \\ \hline \hline
		Ours w/o domain adaptation     & 90.8                   & 68.5               &  54.9               \\ \hline		
		Ours w/ domain adaptation       & \textbf{91.6}                   & \textbf{70.4}              & \textbf{58.0}              \\ \hline
	\end{tabular}
	\vspace{3px}
	\caption{Quantitative results of semantic segmentation over the \textit{Foggy Cityscapes} \cite{cityscapes} test dataset (partially synthetic data) of our approach with and without using domain adaptation.}
	\label{tab:city_res}
	%\vspace{-14px}
\end{table}

With both qualitative and quantitative comparisons against the state-of-the-art approaches, we assess our approach on the aforementioned benchmark (foggy weather conditions datasets). As an initial step, we evaluate the semantic segmentation performance directly on the foggy scenes without \textit{domain adaptation}
. Here, we deal with the semantic segmentation model (shown in Figure \ref{fig:seg-net}) as an independent model and isolated it from the entire pipeline (illustrated in Figure \ref{fig:arch} and including the \textit{domain adaptation} sub-component) to investigate its performance on the foggy dataset. As seen in Table \ref{tab:main_res} and Figure \ref{fig:main_res}, our model fails to produce any desirable improvements in terms of qualitative and quantitative results. However, our approach with \textit{domain adaptation} (Figure \ref{fig:arch}, lower) provides the best results when compared without \textit{domain adaptation} (Table \ref{tab:main_res} and Figure \ref{fig:main_res}). Overall, we consider \textit{domain adaptation} as a necessary step to correct foggy scenes before feeding them into the segmentation network.

\begin{figure*}[t!]
	\begin{minipage}[b]{1.0\linewidth}
		\centering
		\centerline{\includegraphics[width=\columnwidth]{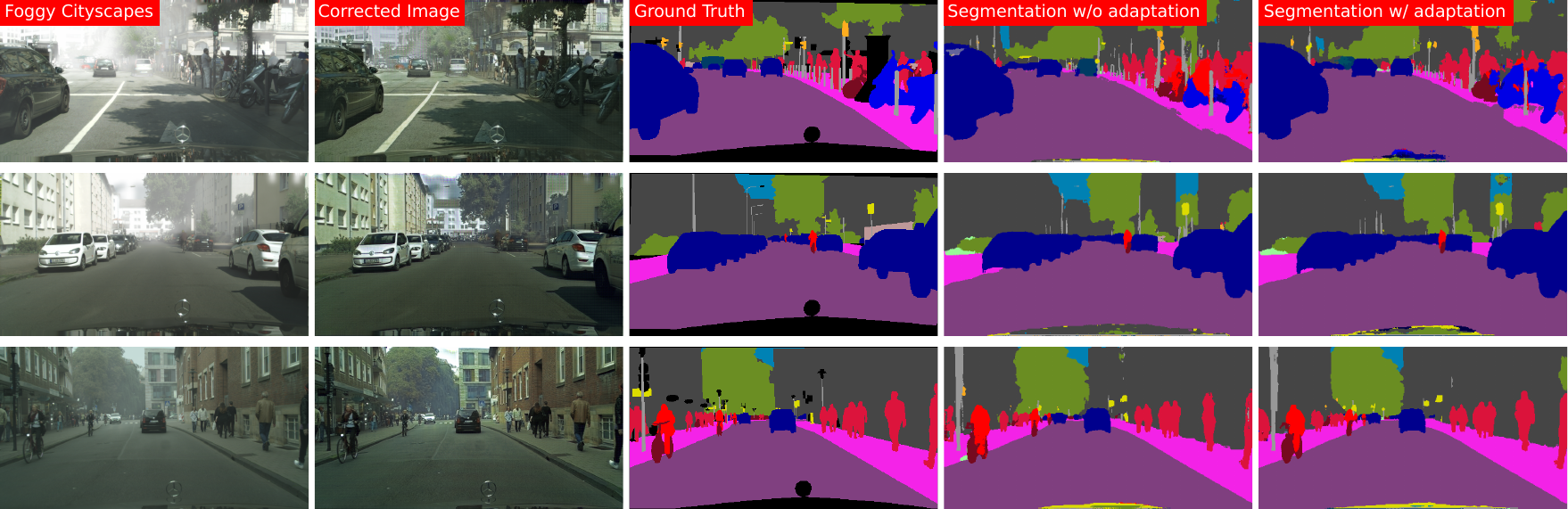}} %myModel3 16.0 w 13.3 h
	\end{minipage}
	\vspace{-.55cm}
	\caption{Semantic segmentation predictions on the \textbf{Foggy Cityscapes} \cite{foggy18} for the proposed approach. The left column shows three scenarios of the dataset followed by \textbf{Corrected Image:} using \textit{domain adaptation}; \textbf{Ground Truth:} ground truth segmentation; \textbf{Segmentation w/o adaptation:} segmentation results without using \textit{domain adaptation}; \textbf{Segmentation w/ adaptation:} segmentation results with \textit{domain adaptation}.}
	\label{fig:foggy_city}
	%\vspace{-14px}
\end{figure*}

As an initial evaluation applied to synthetic data, we test our approach on the \textit{Foggy Cityscapes} \cite{foggy18}. In the evaluation, we consider the test set from the same (\textit{Foggy Cityscapes}) used in training time, which leads to improved segmentation. As seen in the results shown in Table \ref{tab:city_res}, using \textit{domain adaptation} contributes and increases to the mean intersection over union (IoU) by ($3.9\%$) when compared with no domain adaptation. Figure \ref{fig:foggy_city} shows qualitative results on \textit{Foggy Cityscapes} \cite{foggy18}, for our approach with and without using domain adaptation.

Furthermore, we evaluate the performance of scene understanding and segmentation on the real-world datasets, \textit{Foggy Driving} \cite{foggy18} and \textit{Foggy Zurich} \cite{foggy19}, with and without applying \textit{domain adaptation} (\ref{sec:transfer}). This task is a more challenging as our model has not been training on the aforementioned datasets. Without any \textit{domain adaptation}, our approach does not produce any qualitatively or quantitatively desirable results. However, with \textit{domain adaptation} (Section \ref{sec:transfer}), our approach achieves superior results when compared to the absence of \textit{domain adaptation}, in the mean intersection over union (IoU) scores of ($29.4\%$) (\textit{Foggy Driving}) and  ($21.4\%$) (\textit{Foggy Zurich}) (see Table \ref{tab:main_res}). Figure \ref{fig:main_res} shows qualitative results on \textit{Foggy Driving} \cite{foggy18} and \textit{Foggy Zurich} \cite{foggy19} significant differences are evident between the two aforementioned methods. 

As a comparison with the state-of-the-art semantic segmentation under foggy weather conditions, our approach with \textit{domain adaptation} outperforms the work of \cite{foggy_pure} (see Table \ref{tab:main_res}). However, our proposed approach remains competitive with approaches such as \cite{foggy18, foggy19, foggy_old, foggy_pure}, as demonstrated in Table \ref{tab:main_res}. However, all comparators use off-the-shelf segmentation networks such as RefineNet \cite{refineNet17}, DeepLab \cite{deeplab}, Dilated Conv Net \cite{dcn}, and BiSeNet \cite{bisenet}, which offer higher segmentation accuracy due to their use of complex architectures at the expense of viable real-time performance. Using our purpose architecture requires less computational complexity and offers real-time inference performance, which represents an important aspect of our proposed approach. As shown in Table \ref{tab:main_res}, our approach with a significant number of parameters when compared to contemporary state-of-the-art architectures, enables a real-time inference speed of $20$ -- $42$ fps with and without the use of \textit{domain adaptation} respectively, enabling a real-time performance. Further evidence of the efficacy of our approach is being trained on less data (\textit{Foggy} Cityscapes \cite{foggy18}), unlike \cite{foggy_pure} that have been trained on more datasets (\textit{Foggy} Cityscapes \cite{foggy18} and Foggy \textit{Synscapes} \cite{foggy_pure}), which contributes to higher accuracy but at the expense of higher computational complexity.

	\section{Conclusion}
%\label{sec:conclusion}
%\vspace{-9px}
This paper proposes a novel end-to-end automotive semantic segmentation within foggy scene understanding. Using a unified model, we make use of domain adaptation (GAN-based) \cite{CycleGAN2017} to adapt a scene taken in \textit{foggy} weather conditions to \textit{normal} thus increasing the scene visibility. Subsequently, the adapted images are fed to an effective semantic segmentation model for training. For real-time performance, our segmentation network is based on light-weight architecture that includes features fusion, dense connectivity and skip connections, making the approach real-time ($20$ -- $42$ fps with and without \textit{domain adaptation} respectively). As a result, the performance of our approach has progressively improved and achieved significant performance over the state-of-the-art semantic segmentation under foggy weather conditions \cite{foggy18, foggy19, foggy_old}.

	{\small
\bibliographystyle{lib/IEEEtranS}
\bibliography{ref/ref}
} % references
	%\newpage
	%\input{tex/8.end-matter.tex}

% that's all folks
\end{document}